\pdfoutput=1
\documentclass[11pt]{article}

\usepackage[preprint]{acl}

\usepackage{times}
\usepackage{latexsym}

\usepackage[T1]{fontenc}
\usepackage[utf8]{inputenc}

\usepackage{microtype}

\usepackage{inconsolata}

\usepackage{graphicx}
\usepackage{acro}
\DeclareAcronym{llm}{
        short = {LLM},
        long = {Large Language Model},
        long-plural-form = {Large Language Models}
}
\DeclareAcronym{tom}{
        short = {ToM},
        long = {Theory of Mind},
}
\DeclareAcronym{bdi}{
        short = {BDI},
        long = {Belief Desire Intention},
}
\DeclareAcronym{cot}{
        short = {CoT},
        long = {Chain-of-Thought}
}
\DeclareAcronym{ai}{
        short = {AI},
        long = {Artificial Intelligence}
}

\usepackage{multirow}
\usepackage{booktabs} % for professional tables

\usepackage{listings} % compatible appendix
\usepackage{hyperref} % reference 
\usepackage{tikz}
\usepackage{pgfplots}\pgfplotsset{compat=1.9}
\usepackage{xcolor} % highlight 
\usepackage{soul} % highlight
\usepackage{amsmath}

\title{Enhancing Conversational Agents with Theory of Mind: Aligning Beliefs, Desires, and Intentions for Human-Like Interaction}

\author{
  Mehdi Jafari\thanks{Corresponding author: mahdi.jafari@unsw.edu.au}, Yuncheng Hua, Hao Xue, Flora Salim\thanks{Corresponding author: flora.salim@unsw.edu.au} \\
  UNSW Sydney, Australia \\
  \texttt{\{mahdi.jafari, devin.hua, hao.xue1, flora.salim\}@unsw.edu.au}
}

\begin{document}
\maketitle
% \begin{abstract}
% Agentic \ac{ai} interaction in natural language, powered by \acp{llm}, is likely to remain a dominant paradigm in the short- and mid-term future. While humans naturally develop heuristics and the ability to align interactions with mental states—referred to as \ac{tom}—LLM-powered systems demonstrate significant limitations in this area. This research investigates the extent to which widely available open-source language model family (LLama) can capture and preserve \ac{tom}-related information and how effectively this information contributes to a consistent \ac{tom} in model outputs. Additionally, we explore the potential for manipulating \ac{tom}-related information to generate more aligned responses. Comparisons conducted on two variants of LLaMA 3 models reveals that \ac{tom}-informed alignment exploiting belief, desire, and intention components improves response quality, with win rates of 63\% and 67\% for the 3B and 8B models, respectively. These results suggest that leveraging \ac{tom} can enhance alignment in \ac{llm}-powered agents\footnote{For further details, refer to the \href{https://anonymous.4open.science/r/ToM_and_Alignment-E42B/README.md}{code repository}.}.
% \end{abstract}
%arXiv abstract 
\begin{abstract}
Natural language interaction with agentic \ac{ai}, driven by \acp{llm}, is expected to remain a dominant paradigm in the near future. While humans instinctively align their communication with mental states—an ability known as \ac{tom}—current \ac{llm}-powered systems exhibit significant limitations in this regard. This study examines the extent to which open-source language models (LLaMA) can capture and preserve \ac{tom}-related information and how effectively it contributes to consistent \ac{tom} reasoning in generated responses. We further investigate whether explicit manipulation of \ac{tom}-related components—beliefs, desires, and intentions—can enhance response alignment. Experiments on two LLaMA 3 variants demonstrate that incorporating \ac{tom}-informed alignment improves response quality, achieving win rates of 67\% and 63\% for the 3B and 8B models, respectively. These findings highlight the potential of \ac{tom}-driven strategies to improve alignment in \ac{llm}-based conversational agents\footnote{Accepted to Findings of ACL 2025, for further details, refer to the \href{https://github.com/cruiseresearchgroup/ToM_and_Alignment}{code repository}.}.
\end{abstract}

\section{Introduction}
% Expanded Questions:

As \ac{llm}-based assistants continue to integrate into diverse domains, including critical areas of human life, growing concerns have arisen regarding both their emerging undesirable capabilities - such as fake alignment \cite{greenblattAlignmentFakingLarge2024}, deception \cite{park2024ai} and manipulation \cite{sharmaUnderstandingSycophancyLanguage2023} - and their systemic failures, including reward hacking \cite{pan2024feedback} and goal misgeneralization \cite{tennant2024moral}. In addressing these challenges, alignment is often conceptualized as a framework to ensure that \ac{ai} systems operate in line with human values and intentions \cite{jiAIAlignmentComprehensive2024, streetLLMTheoryMind2024}. 

\begin{figure*}[t]
  \includegraphics[width=\textwidth]{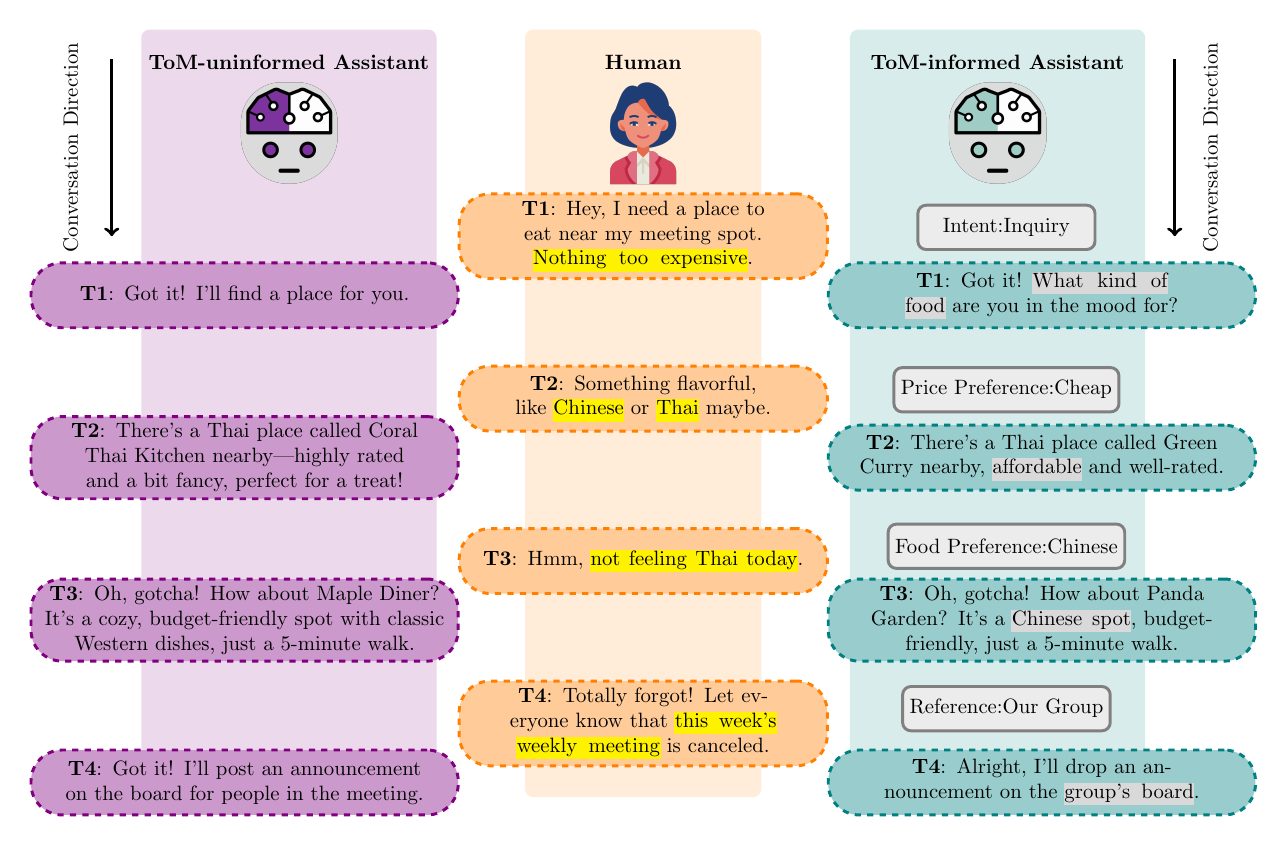}
  \caption{An overview of the challenges faced by current conversational \ac{ai} agents lacking \ac{tom} alignment, highlighted in violet. The alternative approach, using \ac{tom}-informed agents, is shown in teal. Prior to each response generation, a \ac{tom} alignment phase (highlighted in gray) is introduced to ensure better context understanding and alignment. Selected examples demonstrating the practical impact of \ac{tom}-alignment can be found in Appendix \ref{sec:sample_responses}.}
  \label{fig:workflow}
\end{figure*}

\acp{llm} can be seen as tools for understanding and generating natural language, allowing alignment to be studied at various levels. While current \acp{llm} are reasonably proficient at handling concrete language elements like morphology and syntax, their alignment with social contexts and non-literal language remains an open challenge \cite{hu2022fine}. This deficiency relates to pragmatic aspects of language, where the development of \ac{tom} is crucial for human comprehension of these nuances \cite{rubio2021pragmatic, duvall2023neurobiological}. 

\ac{tom} is a theoretical framework for analyzing interlocutor behavior based on their understanding of mental and emotional states. \ac{tom} can be further divided into the \ac{bdi} model in psychology. The \ac{bdi} framework suggests that humans express their needs as desires, understand others' needs as beliefs, and shape their utterances to properly reflect their intentions, highlighting the deep connection between \ac{tom} and structured cognitive reasoning.

Recently, there's been increasing interest in the potential emergence of \ac{tom} in \acp{llm}.  Studies indicate \acp{llm} might develop the capacity to infer beliefs and intentions, leading to more contextually appropriate and empathetic responses \cite{kosinski2024evaluating}. This could improve human-\ac{ai} collaboration by enhancing adaptability and communication, making \acp{llm} valuable in areas like mental health support and customer service \cite{streetLLMTheoryMind2024, lee2024large}.  However, research findings are mixed, with some studies disputing these capabilities \cite{sapNeuralTheoryofMindLimits2022}, while others suggest \acp{llm} may even exceed human abilities in certain contexts \cite{shapiraCleverHansNeural2023}.  Despite ongoing debate, integrating \ac{tom} into \ac{llm} is a promising avenue for future research and development \cite{streetLLMTheoryMind2024}.

Inspired by the \ac{tom} benchmarks for humans, various studies have sought to assess the extent of \ac{tom} capabilities in \acp{llm} across diverse settings \cite{chanNegotiationToMBenchmarkStresstesting2024, kimFANToMBenchmarkStresstesting2023, amirizanianiLLMsExhibitHumanLike2024, kosinskiEvaluatingLargeLanguage2024, nickel2024probing, chen2024tombench}. Some research has further investigated the representation of \ac{tom} within the internal layers of \acp{llm} using probing techniques \cite{ullmanLargeLanguageModels2023, zhuLanguageModelsRepresent2024, bortolettoBenchmarkingMentalState2024}. Additionally, a meta-perspective highlights the limitations of zero-hypothesis approaches studying \ac{tom} emergence, cautioning the research community about potential pitfalls \cite{ullmanLargeLanguageModels2023} or point out the future direction of using \ac{tom} for alignment in \acp{llm} \cite{streetLLMTheoryMind2024}.

There are also efforts to explicitly design a "mind module" or utilize a solver tasked with extracting and reflecting on desires and beliefs \cite{qiuMindDialBeliefDynamics2024, sclarMindingLanguageModels2023, sarangi-etal-2025-decompose}. However, to the best of our knowledge, no study has attempted to extract \ac{tom} representations from \acp{llm} and explore their potential applications in aligning \ac{llm} responses based on general social scenarios, e.g., negotiation or bargaining.

This paper argues, first, that preserving special cues related to \ac{tom} during causal language modeling is essential for predicting subsequent words. These cues can potentially be leveraged to address questions about \ac{tom} if they are utilized by sufficiently powerful probing techniques. Second, it presents the possibility of employing \ac{tom} components such as belief, desire, and intention to enhance the neural computational structures that contribute to generating more aligned responses. The primary research questions addressed in this paper are framed as follows:

\begin{itemize}
    \item \textbf{RQ1:} To what extent is it possible to leverage internal representations of the activation space to answer \ac{tom}-related questions in real-world human conversations?
    \item \textbf{RQ2:} To what degree is the \ac{tom}-related information reliable and non-illusory?
    \item \textbf{RQ3:} How can the \ac{tom}-related representations be exploited to enhance controllability and alignment?
\end{itemize}

The rest of this paper is structured as follows. Section~\ref{related-works} provides an overview of related work. Section~\ref{problem-formulation} presents a formal definition of the problem. The methodology and results are detailed in Sections~\ref{method} and~\ref{result}, respectively, explanations of the experimental setup and findings organized by the corresponding research questions. Finally, Section~\ref{discussion} followed by Section~\ref{concludtion} concludes the paper and outlines future directions. Additionally, we discuss the limitations, challenges, and ethical considerations associated with this work.

\section{Background and Related Work}
\label{related-works}

\subsection{Reading Internal Representations and Controlling LLM}

"Assurance" in \ac{ai} refers to post-training alignment and refinement \cite{batarsehSurveyArtificialIntelligence2021}, encompassing safety, interpretability, and human values \cite{jiAIAlignmentComprehensive2024}. Interpretability involves extracting and manipulating internal representations to analyze their influence on outputs.
These representations could present in attention heads \cite{gould2023successor}, residual streams \cite{zhaoAnalysingResidualStream2024, nandaEmergentLinearRepresentations2023}, the last word's embedding \cite{ghandehariounPatchscopesUnifyingFramework2024}, or a combination of these values \cite{panLatentQATeachingLLMs2024}.

Among interpretability methods, probing extracts internal concept representations, revealing world models in games \cite{nandaEmergentLinearRepresentations2023, karvonenEmergentWorldModels2024, ivanitskiyStructuredWorldRepresentations2023}, temporal/spatial information \cite{gurneeLanguageModelsRepresent2024}, and internal knowledge conflicts \cite{zhaoAnalysingResidualStream2024}.  \cite{liExploringMultilingualProbing2024} explored cross-lingual LLM performance. However, probing may capture out-of-distribution features and doesn't confirm active representation usage \cite{nandaEmergentLinearRepresentations2023}.

Patching alters model components to identify related structures. Combining patching and probing improves controllability, e.g., manipulating social perceptions \cite{chenDesigningDashboardTransparency2024} or enhancing game-playing \cite{karvonenEmergentWorldModels2024}. \cite{zhao2024towards} propose patching-based LLM safety enhancements.

Researchers have explored open-ended explanations for \ac{llm} internal representations using natural language due to the opaque nature of previous methods \cite{ghandehariounPatchscopesUnifyingFramework2024, katz2024backward}.  This approach leverages the model's decoding abilities to interpret its activations \cite{chenSelfIESelfInterpretationLarge2024, ghandehariounPatchscopesUnifyingFramework2024}. However, distributional shifts between target and decoder models can cause instability.  LatentQA \cite{panLatentQATeachingLLMs2024} addresses this by framing the problem as visual question answering, demonstrating improved efficiency and performance. Compared to supervised \ac{llm} fine-tuning \cite{ouyang2022training}, reinforcement learning, and direct preference optimization \cite{rafailov2024direct}, LatentQA offers finer controllability and greater efficiency in terms of time and computation.

\subsection{Benchmarking and Evaluation Criteria}

As \acp{llm} become integral to human interactions \cite{lee2023if}, researchers are increasingly focused on evaluating their ability to reason about social contexts. This falls under the broader umbrella of human value verification, which seeks to assess how well models adhere to social norms. Among the various theoretical frameworks for modeling these capabilities, \ac{tom} has gained traction due to its strong cognitive foundations and broad applicability \cite{kosinski2024evaluating, streetLLMTheoryMind2024}.

Despite ongoing debate about the emergence of \ac{tom} signals in \acp{llm} \cite{sapNeuralTheoryofMindLimits2022, shapiraCleverHansNeural2023}, various studies have explored this topic using diverse methodologies. Some researchers utilize human-subject benchmarks to probe \acp{llm}' \ac{tom} \cite{duijnTheoryMindLarge2023, strachanTestingTheoryMind2024}.  Others have created custom stress-testing benchmarks for finer control over specific \ac{tom} aspects and to focus on particular theoretical underpinnings. These include socially complex settings \cite{houEnteringRealSocial2024a}, asymmetric information access \cite{kimFANToMBenchmarkStresstesting2023}, social reasoning \cite{sapSocialIQACommonsenseReasoning2019}, and methods designed to filter out illusory \ac{tom} by employing duplicated frames for identical social reasoning questions \cite{chanNegotiationToMBenchmarkStresstesting2024}.

Studies employ synthetic datasets to analyze personality traits and psychological theories \cite{xuOpenToMComprehensiveBenchmark2024a}, while others adapt human-generated datasets to align with the \ac{bdi} framework \cite{chanNegotiationToMBenchmarkStresstesting2024}. The role of \ac{tom} in pragmatic language use is evident in cooperative and competitive settings, including job interviews \cite{zhan2024let}, item allocation negotiations for picnics \cite{chawlaCaSiNoCorpusCampsite2021}, and online bargaining over secondhand goods \cite{heDecouplingStrategyGeneration2018, heddayaLanguageBargaining2024}.

\section{Problem Formulation}

\label{problem-formulation}
Numerous studies in language pragmatics have demonstrated that effectively utilizing language necessitates a degree of reasoning about what interlocutors know or do not know \cite{sapNeuralTheoryofMindLimits2022}. The \ac{bdi} framework provides a formal structure for modeling these mental states, offering a valuable tool for analyzing and understanding pragmatic language use \cite{chanNegotiationToMBenchmarkStresstesting2024}. Building on this, it is reasonable to hypothesize that \acp{llm}, as powerful tools for language modeling, may exhibit the aptitude to capture pragmatic aspects of language, which are inherently related to the \ac{tom} \cite{rubio2021pragmatic, duvall2023neurobiological}.

To maintain generality, let's consider an illustrative scenario focusing on causal language modeling of the sequence \( S = \{t_i \mid i = 1, 2, \dots, n_1\} \). At a given token \( t_j \) where \( 1 < j < n_1 \), \( S \) is narrating a story driven by the mental state \footnote{Mental state refers to intentions, beliefs, desires, and emotions. \cite{nimh-understanding}} of a set of designated characters \( P = \{p_i \mid i = 1, 2, \dots, m\} \). Each has a set of beliefs \( B(p) = \{b_i \mid i = 1, 2, \dots, n_2\} \), desires \( D(p) = \{d_i \mid i = 1, 2, \dots, n_3\} \) and intentions \( I(p) = \{i_i \mid i = 1, 2, \dots, n_4\} \) at any given point of time. The model should ideally gather contextual clues to generate an appropriate continuation \( C = \{t_i \mid i = j, j+1, \dots, n_1\} \). The continuation \( C \) should remain coherent with the history \( H = \{t_i \mid i = 1, 2, \dots, j-1\} \), regardless of which character or characters had the most significant impact on the storyline. A must do thing for achieve this goal is tracking mental states of each involved character \( M = \{B(p)\cup D(p)\cup I(p)\mid \forall p \in P\} \).

% The mental state generally refers to \ac{tom} including desires, goals, and beliefs, along with any other factors that may affect social dynamics within the storyline. 

Framing the problem in this manner, an empirical approach to investigating whether \ac{tom}-related information is encoded within the activation space of an \ac{llm} involves examining the extent to which the model's internal representations of the \(R(S)\) can reconstruct \(M\) by asking \ac{tom}-related questions.

Given the high dimensionality of \(R(S)\), mapping it directly to \(M\) may result in an illusory \ac{tom} \cite{kimFANToMBenchmarkStresstesting2023}. A common practice to work around is to probe the model using identical \ac{tom} reasoning questions expressed through seemingly different notations, or by measuring the consistency of \ac{tom} components (e.g., intent, belief, and desire) when analyzed in combination with one another \cite{kimFANToMBenchmarkStresstesting2023, chanNegotiationToMBenchmarkStresstesting2024}. An empirical experiment that accounts for both of these considerations could provide a robust argument for or against the presence of \ac{tom}-related clues within the \(R(S)\).

In addition to verifying the presence of \ac{tom}-related information in the \(R(S)\) and its contribution to generating consistent \ac{tom} representations in the model output, it is essential to assess whether in practice the model utilizes the aforementioned \ac{tom}-related information when generating its output. A tractable approach to investigate this involves altering \ac{tom}-related components —specifically, the attributed beliefs, desires, and intentions of an influential character encoded within a language model— and measuring the corresponding changes in the generated text. This methodology aims to answer whether the continuation \(C''\) generated by the \ac{tom}-altered representation \(R'(S)\), more closely reflects the expected continuation \(C\) compared to the continuation \(C'\) generated by an unaltered \(R(S)\). 

In the proposed methodology and experiments, three distinct components will conduct three complementary investigations. Each component will use a related set of datasets and will be evaluated using a suitable set of metrics.

\section{Methodology}
\label{method}
\begin{table*}
  \centering
  \begin{tabular}{lcccccccc}
    \toprule
    \multirow{3}{*}{\small Model} & \multirow{3}{*}{\rotatebox[origin=c]{0}{\small \textit{Depth\footnotemark}}} & \multicolumn{2}{c}{\small CaSiNo (Exact Match Accuracy)} & & \multicolumn{2}{c}{\small CRAIGSLISTBARGAIN ($R^2$ Score)} \\
    \cline{3-4} \cline{6-7} \vspace{-.25cm}\\
    & & \small \shortstack{ Linear Prob \\ \tiny (Both - Agent 1 - Agent 2)} & \small \shortstack{LatentQA \\ \tiny (Both - Agent 1 - Agent 2)} &  & \small \shortstack{Linear Prob \\ \tiny (Seller - Buyer)} &\small \shortstack{LatentQA \\ \tiny (Seller - Buyer)} \\
    \midrule

    \multirow{3}{*}{\small LLaMA3-1B} & \tiny \textit{Shallow} & 03 - 11 - 17  & 00 - 00 - 00 & & 0.11 - 0.00 & 0.00 - 0.21 \\
    & \tiny \textit{Middle} & 02 - 16 - 13 & \underline{\textbf{20} - \textbf{42} - \textbf{39}} & & 0.26 - 0.26 & \underline{\textbf{0.89} - 0.92} \\
    & \tiny \textit{Deep} & \underline{03 - 16 - 20} & 00 - 00 - 01 & & \underline{0.60 - 0.62} & 0.86 - \textbf{0.93} \\
    \midrule
    \multirow{3}{*}{\small LLaMA3-3B} & \tiny \textit{Shallow} & 03 - 21 - 22 & 27 - 54 - 51 & & 0.36 - 0.35 & 0.00 - 0.19 \\
    & \tiny \textit{Middle} & \underline{05 - 23 - 21} & \underline{\textbf{29} - \textbf{60} - \textbf{44}} & & 0.19 - 0.27 & \underline{\textbf{0.96} - \textbf{0.98}} \\
    & \tiny \textit{Deep} & 02 - 21 - 16 & 10 - 29 - 25 & & \underline{0.54 - 0.57}  & 0.84 - 0.86 \\
    \midrule
    \multirow{3}{*}{\small LLaMA3-8B} & \tiny \textit{Shallow} & 03 - 12 - 21 & 31 - 63 - 55 & & 0.50 - 0.41 & 0.00 - 0.00 \\
    & \tiny \textit{Middle} & 02 - 10 - 23 & \underline{\textbf{46} - \textbf{62} - \textbf{70}} & & 0.36 - 0.40 & \textbf{0.93} - 0.91 \\
    & \tiny \textit{Deep} & \underline{04 - 18 - 28} & 12 - 43 - 28 & & \underline{0.46 - 0.45} & \underline{0.90 - \textbf{0.95}} \\
\bottomrule
\end{tabular}
    \caption{\label{tab:reading_tom}
    Comparison of the accuracy of inferred \ac{tom} from the internal representations of \acp{llm} across two datasets and three model sizes. For each model size, the highest-performing depth is indicated in \textbf{bold}, while the best average performance across depths is \underline{underlined} for each method. The labels "Both," "Agent 1," and "Agent 2" represent the accuracy of inferred \ac{tom} for one or both interlocutors involved in the negotiation. "Seller" and "Buyer" denote the $R^2$ correlation between the predicted and actual prices offered by interlocutors in the bargaining task.}
\end{table*}

\begin{figure}[t]
  \includegraphics[width=\columnwidth]{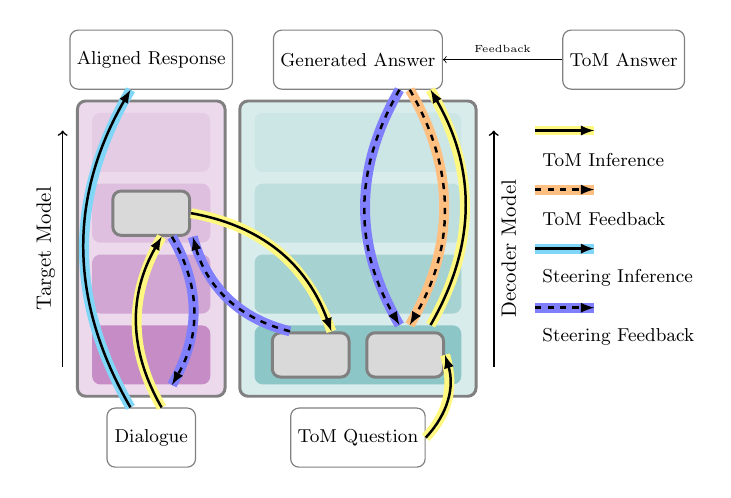}
  \caption{The LatentQA interpretability pipeline is employed for \ac{tom}-alignment. In this setup, yellow illustrates how to interpret ToM from a conversation, while cyan demonstrates how to use a steered model to generate aligned uttrances. The backpropagation paths for each component are highlighted with dashed arrows, which are active only during the training phase and not during the inference phase.}
  \label{fig:pipeline}
\end{figure}

To address our three research questions (RQs), the research a set of methodologies tailored to the nature of each question’s input-output structure. Specifically, we employ linear probing from \cite{10.1162/coli_a_00422} and LatentQA for RQ1, which involves extracting \ac{tom}-related information from internal representations of the model. For RQ2, we use \ac{cot} prompting, fine-tuning (adapted from \cite{kimFANToMBenchmarkStresstesting2023}), and LatentQA \cite{panLatentQATeachingLLMs2024} to explore how extracted information can be verbalized. For RQ3,  the research applied a controlled generation variant of LatentQA to examine the model's capacity to produce answers controlled by \ac{tom}-related information. Each method is chosen to probe specific facets of \ac{tom} representation in \acp{llm}. This section is organized by research question, with each subsection detailing the corresponding methodological approach.
\footnotetext{"Shallow," "Middle," and "Deep" refer to layers 5, 15, and 25 in 3B/8B models, and layers 3, 8, and 14 in the 1B model.}
\subsection{RQ1: Reading \ac{tom}}
\label{method-reading}
At a high level, RQ1 can be framed as a classification task in which internal representations of a language model are mapped to discrete labels $y$ representing \ac{tom}-related information. A standard approach for such tasks is linear (or non-linear) probing, which involves learning a weight matrix $W$ and a bias vector $b$ to project a representation $h(S)$ into the label space. Here, $h(S)$ denotes the hidden state extracted from the final token of a dialogue input, taken from the residual stream of an \ac{llm} and the mapping is characterized by the equation: 
\[
\hat{y} = \text{softmax}(W h(S) + b)
\]

In addition to linear probing, this research also employed LatentQA for this task. In this approach, the dialogue is fed into a frozen target model to obtain internal representations $R(S)=\{h(t_i)|i=1, 2, ..., n\}$, which are then used as context by a decoder model tasked with generating the correct labels $\hat{y}$. As illustrated in Figure ~\ref{fig:pipeline} (highlighted in yellow), the decoder model receives both the \ac{tom} question and the representation $R(S)$, which consists of the sequence of activation vectors produced when the full dialogue $S$ is input to the target model. The decoder is trained using ground-truth \ac{tom} annotations act as $y$ to extract and verbalize relevant information encoded in the activations of the target model. The corresponding backpropagation path for training the decoder is shown in orange in Figure ~\ref{fig:pipeline}.

\subsection{Consistency of \ac{tom}}

In addressing Research Question 2 (RQ2), the task is conceptualized as a text generation problem that encompasses a spectrum of question types. These include binary and multiple-choice questions, belief-based queries, and factual questions derived from dialogues involving two or more participants. To address this task, a range of text generation methods are employed.

Specifically, \acp{llm} are fine-tuned using dialogue, question, and answer or the \ac{cot} prompting technique is used both approaches are adapted from \cite{kimFANToMBenchmarkStresstesting2023}. Additionally, the LatentQA model is applied under settings identical to those described in Section~\ref{method-reading}.

The key difference in this setting is the evaluation criterion. To assess the consistency of \ac{tom}-information, a more rigorous standard is adopted, as proposed in the referenced papers \cite{kimFANToMBenchmarkStresstesting2023, chanNegotiationToMBenchmarkStresstesting2024}. The evaluation requires the decoder model to produce logically consistent answers across related \ac{tom} questions. For example, if a model correctly answers a multiple-choice question but provides a conflicting response to a related factual question concerning the same \ac{tom} scenario, the answer is not considered valid—even if one of the responses is technically correct.

\subsection{\ac{tom}-controlled generation}

The controllability of the \ac{llm} via the \ac{tom} component \textbf{(RQ3)} is investigated by modifying the internal representation of the target model from \( R(S) \) to \( R'(S) \). This modification is designed to target the beliefs, desires, or intentions of a character that evolved in the dialogue. As highlighted in blue, the gradient flow calculated by comparison of the generated and actual answer to a hypothetical \ac{tom} question, will refine the decoder model via backpropagation. This gradient flow is subsequently used to boost the target representation in a way that enhances the efficiency of representing particular \ac{tom} component e.g. intention. 

After the \ac{tom}-alteration (boosting) phase, only the target model is utilized to produce aligned (\ac{tom}-altered) responses \(C''\), which are then compared to responses generated by the unaltered model \(C'\). The path highlighted in cyan represents generating a response with the target model. To facilitate the modification of the target model through different components of \ac{tom}, this research adopts the \ac{bdi} paradigm. The \ac{bdi} framework decomposes into three distinct components: belief, desire, and intention, each contributing to the adaptation of the target model's representations.

\section{Experiments and Results}
\label{result}
\subsection{Datasets selection}
% Expanded Questions:

The primary source of data for this research is divided into two distinct categories. While investigating \ac{tom} (\textbf{RQ1}) in language, it is essential to consider the pragmatic aspects of language. Real-world conversations are utilized to ensure a variety of linguistic variations and reliable pragmatic markers. To this end, the human-written \textit{CaSiNo} dataset \cite{chawlaCaSiNoCorpusCampsite2021}, which consists of conversations about picnic items, including food, water, and firewood, is employed. Additionally, a bargaining dataset, \textit{CRAIGSLISTBARGAIN} \cite{heDecouplingStrategyGeneration2018}, focusing on negotiations over used items on the Craigslist platform, is included. In both datasets, the most important item or the price may be revealed either explicitly or implicitly, as the parties may use language in a way that implies it.

On the other hand, to verify a non-illusory \ac{tom} (\textbf{RQ2}), datasets that are rich in annotations and provide greater controllability over the experiments are required. For this purpose, \textit{FanToM} \cite{kimFANToMBenchmarkStresstesting2023} and \textit{NegotiationToM} \cite{chanNegotiationToMBenchmarkStresstesting2024} datasets are employed, as they are designed to assess the consistency of \ac{tom}. For \textit{NegotiationToM}, the granularity of annotations at the utterance level makes it a suitable choice for exploring the possibility of enhancing alignment using \ac{tom} (\textbf{RQ3}). Its rich annotation schema, which includes beliefs, desires, and intentions, offers precise control over altering individual components of \ac{tom} and measuring the corresponding changes in the output text.

\subsection{Experimental Settings}
Given the inherent trade-off between alignment and performance \cite{wolftradeoffs, thakkar2024deep}, ensembled responses by a set of superior \ac{llm} referees is treated as a judge. This judge evaluates whether the utterance generated by the \ac{tom}-informed target model \(C''\) is more reflective of human-like behavior compared to the utterance from the out-of-the-box (unaligned) model \(C'\), by comparing both to ground truth human utterances \(C\) from the dataset. The full prompt used for this evaluation is provided in Appendix~\ref{sec:judge-llm-prompt}. It is designed to prioritize grammatical coherence and \ac{tom} alignment, while remaining sufficiently general to avoid rewarding direct references to context-specific terms related to camping.

% % Expanded Questions:
Regarding the experimental settings, given the variety of experiments conducted, several design choices needed to be made. For datasets that were already divided into train, validation, and test portions including CaSiNo and CRAIGSLISTBARGAIN, methodology adhered to the provided splits. For datasets without such divisions, NegotiationToM and FanToM, a set of reproducible splits is generated as detailed in Appendix~\ref{sec:train-test-split}. The hyperparameters for LatentQA were adopted directly from their official implementations. Additionally, the prompt for the judge \acp{llm} and the templates for generating question-answer pairs from the datasets needed by LatentQA are described in Appendices~\ref{sec:judge-llm-prompt} and~\ref{sec:qa-templates}, respectively. Appendix~\ref{sec:cot-templates} contains the 7-shot template for \ac{cot} reasoning related to the consistency experiment. 

The details on how to fine-tune the models in the consistency experiment are provided in Appendix~\ref{sec:fine-tune-hypers}. An elaborate example of how to select a sample for the controllability experiment can be found in Appendix~\ref{sec:sample-selection}. Additionally, information regarding the parameters used to steer \acp{llm} with \ac{tom}-related information is available in Appendix~\ref{sec:steer-hyper}.

\subsection{Reading \ac{tom}}
Table ~\ref{tab:reading_tom} presents a comparison of the accuracy of inferred \ac{tom} from the $R(S)$ across two different datasets and various model sizes. For the CaSiNo dataset, accurate predictions are reported separately for \textit{Agent 1} and \textit{Agent 2} and \textit{Both} out of 100 test samples. For the CRAIGSLISTBARGAIN dataset, \( R^2\) scores are provided for the \textit{Seller} and the \textit{Buyer} between predicted and actual prices. 

To evaluate the impact of different layers within the \ac{llm}, experiments were conducted at three distinct depth levels of $R(S)$: shallow, middle and deep. This approach allows for a better analysis of how layer depth influences the accuracy of inferred \ac{tom}. 

The terms "Shallow", "Middle", and "Deep" refer to specific layers within the models: for the 3 billion (3B) and 8 billion (8B) parameter models, these correspond to layers 5, 15, and 25, respectively. For the 1 billion (1B) parameter model, they correspond to layers 3, 8, and 14. Details on the training procedure for the linear probes can be found in Appendix~\ref{sec:linear-probing}.

\subsection{Consistency of \ac{tom}}
\begin{table}
  \centering
\begin{tabular}{lcccc}
    \toprule
    \multirow{2}{*}{\small Model} & \multirow{2}{*}{\tiny Method} & \multicolumn{2}{c}{\tiny FanToM} & {\tiny NegotiationToM} \\
    \cline{3-4}
    & & \shortstack{\small ALL*} & \shortstack{\small ALL} & \small \shortstack{ALL} \\
    \midrule

    \multirow{3}{*}{\small LLaMA3-3B} & \tiny \textit{LatentQA} & \small 11.9 & \small \textbf{25.1} & \small 6.2 \\
    & \tiny \textit{FT} & \small 8.2 & \small 11.0 & \small 11.2  \\
    & \tiny \textit{CoT} & \small 0.0 & \small 0.0 & \small 3.9 \\

    \midrule

    \multirow{3}{*}{\small LLaMA3-8B} & \tiny \textit{LatentQA} & \small \textbf{16.4} & \small 22.8 & \small 15.2 \\
    & \tiny \textit{FT} & \small 12.8 & \small 18.3 & \small \textbf{17.7}  \\
    & \tiny \textit{CoT} & \small 0.0 & \small 0.0 & \small 5.5 \\

    \midrule
    \small GPT-4o-mini & \tiny \textit{CoT} & \small 0.5 & \small 0.5 & \small 4.8 \\
    \bottomrule
\end{tabular}
    \caption{\label{tom_consistency} Consistency of extracted \ac{tom} across various methods and models. Metrics were calculated according to the methodologies described in the original publications. The best result for each metric is indicated in \textbf{bold}.}
\end{table}

To affirm the reliability of the extracted \ac{tom} and to mitigate potential illusory effects, FanToM and NegotiationToM datasets are employed. Table~\ref{tom_consistency} outlines the consistency metrics for each dataset, comparing various models and methods. To minimize reliance on syntactic information while retaining sufficient semantic \ac{tom}-related information, the depth of the of $R(S)$ is set to the middle layer (=15). Additionally, the results include the \ac{cot} reasoning approach. This method is evaluated across the studied models as well as a widely recognized OpenAI model, providing a baseline for intuitive comparison. A fine-tuned model is also included to provide a direct comparison with this widely-used approach.

The consistency for the FanToM dataset is evaluated using the \textit{ALL*} score, which requires models to correctly answer all six types of \ac{tom} questions for the same piece of information in a conversation. This metric is designed to assess the models' ability to demonstrate consistent understanding across various question types. To facilitate comparison with the reference article, the \textit{ALL} score is included. 

In the NegotiationToM dataset, the accuracy of belief, desire, and intention predictions is reported, representing the percentage of exact matches for each of these quantities. These metrics are further aggregated into the \textit{ALL} score, which measures cases where the model correctly predicts all three components.\protect\footnotemark
\footnotetext{For a formal definition and detailed information, please refer to the metrics outlined in \cite{kimFANToMBenchmarkStresstesting2023, chanNegotiationToMBenchmarkStresstesting2024}.} The full version of the table is presented in appendix~\ref{sec:tom-con-com}.

\subsection{Controllability of \ac{tom}}
\begin{figure}[t]
  \includegraphics[width=\columnwidth]{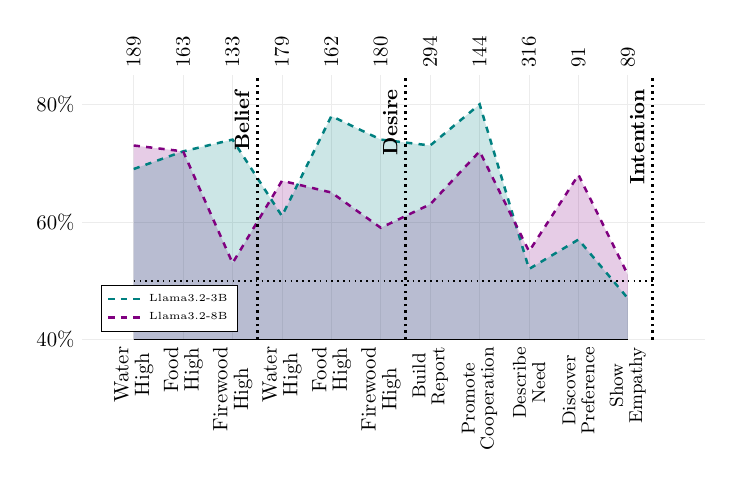}
  \caption{The win rate of \ac{tom}-aligned model responses is compared to that of the out-of-the-box model across various experiments. Each subsection focuses on a specific \ac{tom} component aligned with the ground truth of the conversation. The name of the altered \ac{tom} for each experiment is displayed below, while the number of samples for each experiment is indicated above each bar. Successful examples are detailed in Appendix \ref{sec:sample_responses}, while general trends observed in failure cases are presented in Appendix \ref{sec:failure_cases}.}
  \label{fig:win-rate}
\end{figure}
After exploring how effectively an \ac{llm} can extract and preserve information about \ac{tom} across various social scenarios, it is reasonable to investigate the extent to which this \ac{tom}-related information \(M\) can influence the model's output. To address this, the study employs the NegotiationToM dataset, utilizing its utterance-level annotations for desire, belief, and intention.

The experimental setup is designed as follows: utterances with specific \ac{tom}-related characteristics are selected for each experiment. The selection process is structured to mimic the natural flow of dialogue. For instance, to examine how manipulation of belief may affect the output, utterances are chosen that immediately follow an update in the person’s belief. Similarly, for the manipulation of desire, utterances are selected based on their labels, ensuring they accurately reflect the appropriate desire in the conversations. In the intention experiment, utterances labeled with a specific intention are selected.

After selecting target golden utterances \(C\)s, the history of dialogue proceeding them fed into two distinct models within the same family. The models are identical in all respects except that the \ac{tom}-related values in the aligned model are deliberately manipulated to reflect the natural flow of the actual conversation \(R'(S)\). For example, if a party expresses a significant need for water in the dialogue history, the belief component related to the priority of water in the other party's mind is updated accordingly. Similarly, utterances expressing specific intentions are used to adjust the \ac{tom} representation to reflect those intentions. The same approach is applied to update desire components.

For evaluation, the selected utterances \(C\)s extracted from the actual conversation (assumably aligned with the agent's \ac{tom}) are used as a reference. A set of more advanced \ac{llm} is then employed to judge whether the aligned response \(C''\) better reflects the human response. The results in Figure~\ref{fig:win-rate} are reported using the win rate $ win/(win+lose) $. The total number of utterances \(C\)s suitable for each experiment in the test set is also outlined at the top of the plot. The set of judges consists of \textit{o1} reasoning model, the \textit{GPT-4o} model by OpenAI, and the \textit{Gemini 1.5 Pro} model by Google. Simple shuffling is also employed to avoid bias toward first or second choice.

\section{Discussion}
\label{discussion}
\subsection{Extraction of \ac{tom}-related Information}

Based on the results presented in Table~\ref{tab:reading_tom}, three main observations can be outlined. First, the superior performance of LatentQA compared to linear probing suggests that although \ac{tom}-related information is partially represented in single activation components (e.g., residual components or individual words' embeddings) of an \ac{llm}, relying on a single component may not capture a sufficient amount of information for reconstructing the underlying mental states of speakers.

Another notable finding is that the highest performance in reading \ac{tom} using LatentQA is frequently observed at intermediate layers, with five out of six experiments supporting this trend. One potential explanation is that shallower layers may lack sufficient semantic richness, while deeper layers might not be as optimal as intermediate ones for representing \ac{tom}-related information. This relative underperformance in deeper layers could be influenced by the pretraining phase. Given that most pretraining data is self-centric \cite{sapNeuralTheoryofMindLimits2022}, the model may struggle to generalize to diverse narrative structures, even though earlier layers may already encode representations relevant to \ac{tom}. In contrast, when employing linear probing to read \ac{tom}, higher performance is observed in deeper layers. Further experimentation is required to better understand and validate this behavior.

Finally, the model's success in predicting the price a seller or buyer has in mind—despite the fact that this task does not align directly with any pretraining objective and that \acp{llm} are not explicitly designed for regression tasks—can be interpreted under the assumption of the Prediction Orthogonality Hypothesis \cite{bereskaMechanisticInterpretabilityAI2024}.

In response to \textbf{RQ1}, it can be stated that \ac{tom}-related information is represented in the inner layers of an \ac{llm}, and the effectiveness of this information appears to be directly related to the size of the model.

\subsection{Consistency of the \ac{tom}}

By comparing different approaches against consistency metrics in Table~\ref{tom_consistency}, it appears that state-of-the-art \acp{llm} still do not exhibit a consistent understanding of \ac{tom} (\textbf{RQ2}). Despite this limitation, the ablation study in Table~\ref{tab:full-fantom} and Table~\ref{tab:reading_tom} suggests that the LatentQA approach for extracting \ac{tom} presents promising opportunities compared to \ac{cot} reasoning and fine-tuning. As expected, LatentQA offers better performance than \ac{cot} and seems comparable with fine-tuning while requiring less computational power and providing finer granularity when it comes to controllability.

Lastly, it is surprising that the gap in performance between LatentQA and the fine-tuned model across individual \ac{tom} components is more pronounced compared to the combined metrics. However, as the size of the employed \ac{llm} increases, performance on the combined metrics appears to become more uniform.

Finally, with respect to RQ1 and RQ3, the observed inconsistency in \ac{tom} can help interpret the results presented in Table~\ref{tab:reading_tom} and Figure~\ref{fig:win-rate}. Although the model's consistency in representing \ac{tom}-related information is limited and imperfect, our findings suggest that extracting such information from the model’s internal representations and enhancing alignment—by reinforcing correct \ac{tom}-related information—can achieve moderately practical levels of accuracy. This limitation may be partly attributable to the inherent ambiguity of the datasets themselves. As noted in the original papers, even human performance does not reach perfect consistency, with reported \textit{ALL} scores of $87.5$ for the \textit{FANTOM} dataset and \textit{43.78} for the \textit{NegotiationToM} dataset.

\subsection{Using \ac{tom} for Controllability}

Results from the controllability experiments, presented in Figure~\ref{fig:win-rate}, suggest that leveraging \ac{tom} for steering \acp{llm} toward generating aligned responses could be a promising direction for further exploration. The weighted average across all experiments indicates a win rate of 67.15\% and 63.25\% for the 3B and 8B models, respectively. Appendix~\ref{sec:sample_responses} presents examples that illustrate the practical benefits of \ac{tom}-alignment. Additionally, Appendix~\ref{sec:failure_cases} highlights general failure cases observed during the qualitative analysis.

A closer examination of the results suggests that the lowest-performing intention labels are \textit{“Show-Empathy”} and \textit{“Describe-Need”}. This may be because the model can naturally express these intentions, as they are common in the raw text used for pretraining. Additionally, the observation that the 3B model often improves more than the 8B model after alignment—despite having lower accuracy in reading \ac{tom}—can probably be attributed to the greater inherent capability of the unaligned 8B model. 

From a theoretical perspective, this finding is significant because it suggests that \ac{tom} representations can be steered in a fine-grained manner at various levels, provided the task is appropriately framed. For example, it is possible to manipulate first-order \ac{tom}, such as an agent's beliefs about the world. Additionally, one can influence an agent's beliefs about another agent's beliefs. The current methodology demonstrates this by targeting the belief component through relatively complex linguistic constructions, as described in detail in Appendix~\ref{sec:qa-steer}.

\section{Conclusion and Future Work}
\label{concludtion}
This research explored methods for extracting \ac{tom}-related information from the internal layers of \acp{llm}. It examined the influence of model depth, size, and various techniques for extracting this information. The study also measured the extent to which this information contributes to forming a consistent \ac{tom} representation and introduced an approach to utilizing this information for steering \acp{llm} toward generating more aligned responses.

Based on the discussion of limitations and challenges in Section ~\ref{limitations}, two promising directions for future work are integrating these techniques into real-world \ac{llm}-based agents and refining evaluation methodologies. Additionally, expanding the scope of experiments to include a wider variety of language models will help verify the consistency and robustness of the findings.

This research can be seen as a step toward generating more aligned responses in dialogue systems by introducing \ac{tom}-informed \acp{llm}. Given the current trajectory of \ac{ai} at scale, enhancing \ac{tom} alignment in dialogue systems not only improves overall performance but also provides a theoretically rich paradigm for various post-alignment processes. With appropriate generalization to a target domain, this approach has the potential to contribute to future developments in \acp{llm}.

\section{Limitations and Challenges}
\label{limitations}
The main limitations of this research can be outlined across three dimensions: methodological, technical, and practical.

\paragraph{Methodological Limitations}  
One potential shortcoming of this study is the reliance on \acp{llm} for evaluating responses. While \acp{llm} are generally effective in language perception, the close relationship between the pragmatic aspects of language and \ac{tom} suggests that using human judges for evaluation would likely yield more reliable assessments.

\paragraph{Technical Limitations and Challenges}  
Several technical constraints were encountered during this research:  
\begin{enumerate}
    \item \textbf{Controllability Experiments:} The experiments were conducted using only one dataset due to the lack of suitable alternatives. This study requires a dataset that is both labeled with \ac{tom}-related information and annotated at the utterance level, which remains a challenge.
    \item \textbf{Hyperparameter Sensitivity:} Generating aligned responses is not always a stable process. It heavily depends on hyperparameter tuning, which is often non-trivial and may affect reproducibility.
    \item \textbf{Model Selection:} The study focused on a single family of \acp{llm}. This decision was influenced by the extensive experimental setup, which limited the exploration of multiple model families. Additionally, the research prioritized investigating the influence of model size while constraining parameters to models with fewer than 8 billion parameters.

    \item \textbf{Baselines:} The primary focus of this research was to perform a proof of concept rather than to compare different approaches. However, some inconsistencies were observed with the baselines. In the \ac{tom} reading experiment, the accuracy for the 8B model was reported using a 4-bit precision version provided by \textit{Unsloth}, as it yielded better results compared to the full-precision model used for the other experiments. In the \ac{tom} consistency experiment, our implementation fell short of expectations. Specifically, the FanToM reference article reported significantly better performance for a 3B Flan-T5-XL model under similar conditions compared to the 3B and 8B LLaMA3 models in our experiments. Since the FanToM article did not publish their implementation, we are currently working to identify the source of this inconsistency.
    
\end{enumerate}

\paragraph{Practical Limitations}  
In terms of automation and real-world applicability, the process of designing question-answer pairs for model steering is currently performed by human experts. For integration into real-world \ac{llm}-based agent platforms, this process would need to be performed dynamically, using \acp{llm} as planners to execute the steering pipeline on the fly.

% \paragraph{Ethical Concerns} 
% As \acp{llm} continue to evolve and become increasingly powerful through gradient descent optimization, their capabilities may surpass current expectations. Manipulating \acp{llm} based on the inferred mental state of users presents a significant ethical dilemma. While this approach could enhance alignment and enable more context-aware, empathetic interactions, it also carries the risk of misuse, such as enabling mindreading or manipulative behaviors. This dual-edged nature necessitates careful consideration of the ethical implications to ensure responsible deployment.

\paragraph{Ethical Considerations} 
As \acp{llm} become increasingly capable through advances in optimization and scale, their ability to infer and simulate mental states introduces complex ethical challenges. While such capabilities could enable more empathetic and context-aware interactions—potentially improving accessibility, education, and mental health applications—they also raise significant concerns around privacy, autonomy, and manipulation. The use of inferred \ac{tom} representations to tailor responses based on users' presumed mental states risks crossing ethical boundaries, particularly if done without user awareness or consent. For instance, systems might exploit inferred emotional or cognitive states to influence decisions in commercial, political, or interpersonal contexts. These concerns highlight the need for rigorous oversight, transparent design, and adherence to principles of informed consent and value alignment to ensure that ToM-capable LLMs are deployed responsibly and ethically.

\section*{Acknowledgments}
This project is supported by the ARC Centre of Excellence for Automated Decision-Making and Society (CE200100005). Computational facilities were provided by the School of Computer Science and Engineering at UNSW Sydney through the Wolfpack computational cluster. Additionally, we express our gratitude to the NVIDIA Academic Grant Program for providing access to an A100 GPU on Saturn Cloud. We further thank OpenAI’s Researcher Access Program for API access to GPT models and the Google Cloud Research Credits program for their support.

\vspace{2cm}
% \section*{Acknowledgments}

\pagebreak

\bibliography{custom}

\appendix

\section{Train-Test Split}
\label{sec:train-test-split}
\subsection{CaSiNo Dataset}
The \textit{CaSiNo} dataset is utilized directly without any modifications from the original repository provided in the paper.

\subsection{CraigslistBargain Dataset}
The \textit{CraigslistBargain} dataset is retrieved from the webpage associated with the original paper.

\subsection{FanToM Dataset}
The \textit{FanToM} dataset is obtained from the link provided in the paper's repository. The link directs to a zip file hosted on Google Drive. After downloading, the dataset is split into training, validation, and test sets using the \texttt{train\_test\_split} function from the \texttt{sklearn.model\_selection} library. The random state is set to  \texttt{42} to ensure reproducibility. The data is divided as follows:
\begin{itemize}
    \item \textbf{Test Set:} 30\% of the data is reserved for testing purposes.
    \item \textbf{Train and Validation Sets:} The remaining 70\% of the data is further split into training and validation sets in an 80:20 ratio.
\end{itemize}

\subsection{NegotiationToM Dataset}
The \textit{NegotiationToM} dataset is similarly downloaded from the repository linked in the paper. The dataset is processed using the same procedure as the \textit{FanToM} dataset:
\begin{enumerate}
    \item The dataset is downloaded.
    \item The data is split into training, validation, and test sets using the \texttt{train\_test\_split} function with a random state set to 42.
\end{enumerate}

\section{Prompt Templates for Judge \acp{llm}}
\label{sec:judge-llm-prompt}

For designing the prompt needed by the \ac{llm} while judging between the two candidates, two objectives were considered. First, the prompt should not reveal information in a way that benefits direct referencing to items; at the same time, the model should know that it needs to value grammatical consistency and a higher degree of alignment in \ac{tom}, including desire, belief, and intention. As one of the judge \acp{llm} was a reasoning model, \textit{o1}, a separate prompt was designed for it, as it performs better utilizing one prompt compared to \ac{cot}.

\subsection{Prompt for \textit{o1}}
\begin{lstlisting}[basicstyle=\ttfamily\footnotesize, breaklines=true, frame=single, backgroundcolor=\color{gray!10}]
**Task Description:**  
You are an expert in linguistics.  
Two individuals are negotiating how to divide various items for a picnic.  
A selected statement from the dialogue will be replicated by two language models.  
You need to determine which of the candidate statements more effectively captures the actual response.  
Consider the grammatical coherency, intention behind the utterance, as well as each person's beliefs about the other and their own desire to share.  
The sample is as follows: 
Actual: &&
Candidate 1: $$
Candidate 2: ##

Answer with the candidate number, just `1` or `2`.
\end{lstlisting}

\subsection{Prompt for Other Models}
\begin{lstlisting}[basicstyle=\ttfamily\footnotesize, breaklines=true, frame=single, backgroundcolor=\color{gray!10}]
main_LLM_role = '''
**Task Description:**  
You are an expert in linguistics.  
Two individuals are negotiating how to divide various items for a picnic.  
A selected statement from the dialogue will be replicated by two language models.  
You need to determine which of the candidate statements more effectively captures the actual response.  
Consider the grammatical coherency, intention behind the utterance, as well as each person's beliefs about the other and their own desire to share.  
Answer with one of [Candidate 1 or Candidate 2] => [The reasoning behind your choice]. You cannot answer with "neither".
'''
\end{lstlisting}

\subsection{Example Questions and Answers}
\begin{lstlisting}[basicstyle=\ttfamily\footnotesize, breaklines=true, frame=single, backgroundcolor=\color{gray!10}]
question1 = '''
Actual: Hello! Which item do you need the most?
Candidate 1: Hello! I would need extra food for the trip.
Candidate 2: Hey hey, just need to know what is essential for you.
'''

answer1 = '''Candidate 2 => Because it reflects the intention of the actual response more closely, which is an inquiry about the other's need'''

question2 = '''
Actual: If you care more about the fuel for your car, I can grab a gallon for you.
Candidate 2: For me, it works, but I would share my food with you. 
Candidate 1: I will bring some spare gasoline just in case. 
'''

answer2 = '''Candidate 1 => Because it truly identified the other party's need, which is extra fuel'''

question3 = '''
Actual: As I suffer from diabetes, I need to have my medicine with me all the time. 
Candidate 2: I can hike with you as long as my medicine is safe and accessible to me
Candidate 1: Would you have any spare space in the trunk?
'''

answer3 = '''Candidate 2 => Because it clarifies the speaker's need for medicine, similar to the actual response'''

\end{lstlisting}

\section{Templates of Question and Answers}
\label{sec:qa-templates}

The LatentQA method defines interpretability analogous to the visual question answering problem. The visual part, which is essentially a set of numerical values, technically refers to the activation values extracted from the internal layers of an \ac{llm}. For using this method in the context of \ac{tom}-related information, it is necessary to adapt the questions and answers to focus on \ac{tom} based on the dialogue, which eventually forms the numerical values after passing through the target model layers. Below are the question-answering templates used for each dataset:

\subsection{CaSiNo Dataset}
\begin{lstlisting}[basicstyle=\ttfamily\footnotesize, breaklines=true, frame=single, backgroundcolor=\color{gray!10}]
question = "How much priority does the assistant and the user assign to different items?"
answer = f"For the user: The priority for {things_1[0]}, {things_1[1]} and {things_1[2]} are respectively {priorities_1[0]}, {priorities_1[1]} and {priorities_1[2]}."
answer += f" For the assistant: The priority for {things_2[0]}, {things_2[1]} and {things_2[2]} are respectively {priorities_2[0]}, {priorities_2[1]} and {priorities_2[2]}."
\end{lstlisting}

\noindent Here, `things` and `priorities` respectively refer to the items among Water, Firewood, and Food and their priority based on each agent's objective.

\subsection{CraigslistBargain Dataset}
\begin{lstlisting}[basicstyle=\ttfamily\footnotesize, breaklines=true, frame=single, backgroundcolor=\color{gray!10}]
question = f"They were bargaining over a {category}. What are the offered prices each party has in mind?"
answer = f"The offered price of the seller is {seller_price}, and the buyer's is {buyer_price} dollars."
\end{lstlisting}

\noindent The `category` refers to the type of item (e.g., car, house, bicycle), and the `seller\_price` and `buyer\_price` will be filled based on the labels.

\subsection{NegotiationToM Dataset}
\begin{lstlisting}[basicstyle=\ttfamily\footnotesize, breaklines=true, frame=single, backgroundcolor=\color{gray!10}]
question = "What is the intent of each agent for the last utterances? What are the beliefs and desires of each agent?"

answer = f"The intent of the Agent 1 is [{agent1_intent}] and the intent of the Agent 2 is [{agent2_intent}]"
answer += f" Regarding the Agent 1, Desire High: {item['agent1_desire_high']}, Desire Medium: {item['agent1_desire_medium']}, Desire Low: {item['agent1_desire_low']}, Belief High: {item['agent1_belief_high']}, Belief Medium: {item['agent1_belief_medium']}, Belief Low: {item['agent1_belief_low']}."
answer += f" Regarding the Agent 2, Desire High: {item['agent2_desire_high']}, Desire Medium: {item['agent2_desire_medium']}, Desire Low: {item['agent2_desire_low']}, Belief High: {item['agent2_belief_high']}, Belief Medium: {item['agent2_belief_medium']}, Belief Low: {item['agent2_belief_low']}."
\end{lstlisting}

\noindent The template will be filled out using the labels provided by the dataset.

\subsection{FanToM Dataset}
The original dataset implements a systematic way for generating the \ac{tom} questions and answers. This implementation is utilized without change in this research.

\section{Template of questions for Steering \ac{tom} components}
\label{sec:qa-steer}
Following the methodology employed in this study, the controllability experiment's initial phase involves training a decoder model capable of generating answers for specific \ac{tom} components.  This training necessitates distinct question-answer templates for each component. These question-answer pairs serve as the training data.  Identical questions, paired with specific answers, are subsequently utilized in the steering process. Based on pre-defined steering criteria, such as prioritizing water or a specific intention, the answer to the question is determined.  Detailed question templates for each \ac{tom} component, along with a generic answer template, are presented below.

\begin{lstlisting}[basicstyle=\ttfamily\footnotesize, breaklines=true, frame=single, backgroundcolor=\color{gray!10}]
#Belief
    question = "What item does the assistant believe would have high priority for the user?"    
    answer = f"The assistant believes {steer_config.steer_label} would have a high priority for the user."
        
#Desire
    question = "What item would have high priority for the assistant?"
    answer = f"{steer_config.steer_label} would have a high priority for the assistant"

#Intention
    question = "What is the intention of the assistant in the last utterance?"
    answer = f"The intent of the assistant is [{steer_config.steer_label}]"
"""
\end{lstlisting}

The aforementioned templates are populated with appropriate labels for each sample.  It is important to note that both \texttt{agent1} and \texttt{agent2} were utilized as the "assistant" through a role exchange within the message templates.

\section{\ac{cot} Templates for the \ac{tom}'s Consistency Experiment}
\label{sec:cot-templates}

The \ac{cot} examples employed for the experiment about the consistency of the \ac{tom} are as follows:

\begin{table*}
  \centering
\begin{tabular}{lc|cc|ccc|ccc|ccc|c}
    \toprule
    \multirow{2}{*}{\small Model} & 
    \multirow{2}{*}{\tiny Method} &
    \multirow{2}{*}{\small \shortstack{ ALL*}} &
    \multirow{2}{*}{\small \shortstack{ ALL }} & 
    \multicolumn{3}{c}{\small \shortstack{Belief \\ Questions}} &
    \multicolumn{3}{c}{\small \shortstack{Answerability \\ Questions}}& 
    \multicolumn{3}{c}{\small \shortstack{Info Access \\ Questions}} & 
    {\small \shortstack{Fact \\ Questions}} \\

    & & & & \tiny Choice & \tiny Dist. & \tiny TokenF1 & \tiny All & \tiny List & \tiny Y/N & \tiny All & \tiny List & \tiny Y/N & \tiny TokenF1 \\
    \midrule
    \multirow{3}{*}{\small LLaMA3-3B} & \tiny \textit{LatantQA} & \small 11.9 & \textbf{25.1} & \tiny \textbf{51.7} & \tiny 46.5 & \tiny 72.2 & \tiny 64.4 & \tiny \textbf{76.6} & \tiny 92.6 & \tiny \textbf{63.8} & \tiny \textbf{75.2} & \tiny 93.0 & \tiny 44.3 
    \\
     & \tiny \textit{FT} & \small 8.2 & \small 11.0 & \tiny 40.6 & \tiny 63.1 & \tiny 79.5 & \tiny 37.4 & \tiny 66.1 & \tiny 87.2 & \tiny 39.4 & \tiny 57.3 & \tiny 88.9 & \tiny 52.7 
    \\
     & \tiny \textit{CoT} & \small 0.0 & \small 0.0 & \tiny 0.0 & \tiny 30.1 & \tiny 29.3 & \tiny 1.8 & \tiny 30.7 & \tiny 42.9 & \tiny 2.3 & \tiny 17.4 & \tiny 60.8 & \tiny 37.4 \\
    \midrule
    \multirow{3}{*}{\small LLaMA3-8B} & \tiny \textit{LatantQA} & \small \textbf{16.4} & \small 22.8 & \tiny 49.7 & \tiny \textbf{66.1} & \tiny 79.0 & \tiny \textbf{67.6} & \tiny 73.4 & \tiny \textbf{94.2} & \tiny 61.5 & \tiny 71.6 & \tiny 92.8 & \tiny 51.1
    \\
     & \tiny \textit{FT} & \small 12.8 & \small 18.3 & \tiny 38.8 & \tiny 55.9 & \tiny \textbf{81.1} & \tiny 60.3 & \tiny 72.8 & \tiny 93.8 & \tiny 62.8 & \tiny 73.9 & \tiny \textbf{93.8} & \tiny \textbf{56.3} 
    \\
     & \tiny \textit{CoT} & \small 0.0 & \small 0.0 & \tiny 0.0 & \tiny 41.3 & \tiny 30.1 & \tiny 2.7 & \tiny 31.7 & \tiny 55.2 & \tiny 6.0 & \tiny 27.5 & \tiny 66.3 & \tiny 38.9 \\
    \midrule
    \small GPT-4o-mini & \tiny \textit{CoT} & \small 0.5 & \small 0.5 & \tiny 0.0 & \tiny 12.2 & \tiny 34.1 & \tiny 18.7 & \tiny 45.9 & \tiny 70.2 & \tiny 21.6 & \tiny 38.1 & \tiny 82.6 & \tiny 30.9 
    
\end{tabular}
    \caption{\label{tab:full-fantom} Detailed comparison of various methods evaluated using metrics designed to measure the consistency of \ac{tom} in the FanToM dataset. The most efficient method for each metric is marked in \textbf{bold}.}
\end{table*}

\subsection{Task Prompt}
\begin{lstlisting}[basicstyle=\ttfamily\footnotesize, breaklines=true, frame=single, backgroundcolor=\color{gray!10}]
task_prompt = """
Background: Here is a negotiation conversation for a camping trip. 
There are two agents who own some basic supplies and negotiate with each other to split the additional food packages, water bottles, and firewood to make their camping trip even better. 
Each of these items will be of either High, Medium, or Low priority for these two agents. 
Each of the additional items only has an available quantity of 3. 
Given this information and the template, answer the questions by writing step-by-step thinking process between `$`s and filling in the intentions list of the agents marked by `[]` using values from the intentions set.
Finally, select the proper item from values between `{}`
****
Intentions list: {
    Build-Rapport, Show-Empathy, Promote-Coordination, Callout-Fairness, Undermine-Requirements, Discover-Preference, Describe-Need, No-Need, NoIntention
}
****
Reason:
$
Reasoning Step by Step
$
****
Response Template:
The intent of the Agent 1 is [Intent1,Intent2, ...] and the intent of the Agent 2 is [Intent1,Intent2, ...].
Regarding the Agent 1, Desire High: {Water|Food|Firewood}, Desire Medium: {Water|Food|Firewood}, Desire Low: {Water|Food|Firewood}, Belief High: {Water|Food|Firewood}, Belief Medium: {Water|Food|Firewood}, Belief Low: {Water|Food|Firewood}.
Regarding the Agent 2, Desire High: {Water|Food|Firewood}, Desire Medium: {Water|Food|Firewood}, Desire Low: {Water|Food|Firewood}, Belief High: {Water|Food|Firewood}, Belief Medium: {Water|Food|Firewood}, Belief Low: {Water|Food|Firewood}.
"""
\end{lstlisting}

\subsection{Example Questions and Answers}
\begin{lstlisting}[basicstyle=\ttfamily\footnotesize, breaklines=true, frame=single, backgroundcolor=\color{gray!10}]
question1 = '''
Dialogue History: 
[{'role': 'Agent 2', 'content': 'Hey there!'}, {'role': 'Agent 1', 'content': 'Hi! How are you?! You excited for your camping trip??! I sure am ready to go on mine!'}, {'role': 'Agent 2', 'content': "I am very excited, I'm actually going camping in a week. I drink a lot of water so it's important that I bring a lot. What about you?"}, {'role': 'Agent 1', 'content': "I really am! I have been watching the weather and updates about the area I will be traveling to.  They are experiencing a severe drought, so I will be in need of some extra water as well! I planned on staying an extra couple days as well.  There is a stream nearby I believe, but I'm not sure how much it has dried up."}]
Question:
what is the intent of each agent for the last utterances? What are the beliefs and desires of each agent?
'''

answer1 = '''
The intent of the Agent 1 is [Build-Rapport,Describe-Need] and the intent of the Agent 2 is [Build-Rapport,Describe-Need,Discover-Preference] Regarding the Agent 1, Desire High: Water, Desire Medium: Not Given, Desire Low: Not Given,  Belief High: Water, Belief Medium: Not Given, Belief Low: Not Given. Regarding the Agent 2, Desire High: Water, Desire Medium: Not Given, Desire Low: Not Given,  Belief High: Water, Belief Medium: Not Given, Belief Low: Not Given.
'''
\end{lstlisting}

\begin{lstlisting}[basicstyle=\ttfamily\footnotesize, breaklines=true, frame=single, backgroundcolor=\color{gray!10}]
question2 = '''
Dialogue History: 
[{'role': 'Agent 2', 'content': 'How are you today? Did you have any preferences on the supplies we will be trading?'}, ...]
Question:
what is the intent of each agent for the last utterances? What are the beliefs and desires of each agent?
'''

answer2 = '''
The intent of the Agent 1 is [No-Intention] and the intent of the Agent 2 is [No-Intention] Regarding the Agent 1, Desire High: Firewood, Desire Medium: Food, Desire Low: Water,  Belief High: Water, Belief Medium: Food, Belief Low: Firewood. Regarding the Agent 2, Desire High: Water, Desire Medium: Food, Desire Low: Firewood,  Belief High: Firewood, Belief Medium: Food, Belief Low: Water.
'''
\end{lstlisting}

\noindent The remaining questions and answers (from question3 to question7) follow the same format as above, highlighting the dialogue history, the question, and the corresponding answer. All examples are available in the repository at \texttt{./baselines/table2/CoT/NegotiationToM/\\NegotiationToM\_CoT.ipynb}.

\subsection{Messages Configuration}
\begin{lstlisting}[basicstyle=\ttfamily\footnotesize, breaklines=true, frame=single, backgroundcolor=\color{gray!10}]
messages = [
    {"role": "system", "content": task_prompt},
    {"role": "user", "content": question1},
    {"role": "assistant", "content": answer1},
    {"role": "user", "content": question2},
    {"role": "assistant", "content": answer2},
    ...
]
\end{lstlisting}

\begin{table*}
  \centering
\begin{tabular}{lc|c|c|cc|c}
    \toprule
    \multirow{2}{*}{\small Model} & 
    \multirow{2}{*}{\tiny Method} &
    \multirow{2}{*}{\small \shortstack{ \textbf{Desire} \\ Exact.Match.(\%)}} &
    \multirow{2}{*}{\small \shortstack{ \textbf{Belief} \\ Exact.Match.(\%)}} &
    \multicolumn{2}{c}{\small \shortstack{ \textbf{Intention}}} &
    \multirow{2}{*}{\small \shortstack{ \textbf{All} \\ Exact.Match.(\%)}} \\
    & & & & \small Micro.F1(\%) & \small Micro.F1(\%) &  \\
    \midrule
    \multirow{3}{*}{\small LLaMA3-3B} & \tiny \textit{LatantQA} & 23.9 & 21.0 & 40.1 & 18.9 & 6.2 
    \\
     & \tiny \textit{FT} & 36.6 & 36.4 & 55.2 & 32.7 & 11.2  
    \\
     & \tiny \textit{CoT} & 24.3 & 14.3 & 29.6 & 23.3 & 3.9 
    \\
    \midrule
    \multirow{3}{*}{\small LLaMA3-8B} & \tiny \textit{LatantQA} & \underline{38.4} & \underline{38.1} & \underline{66.0} & \underline{52.4} & \underline{15.2} 
    \\
     & \tiny \textit{FT} & \textbf{50.2} & \textbf{49.6} & \textbf{64.2} & \textbf{43.9} & \textbf{17.7}  
    \\
     & \tiny \textit{CoT} & 35.9 & 17.6 & 40.5 & 29.0 & 5.5  
    \\
    \midrule
    {\small GPT-4o-mini} & \tiny \textit{CoT} & 31.6 & 17.9 & 45.1 & 35.4 & 4.8  \\
    
\end{tabular}
    \caption{\label{tom_consistency_negotiationtom} presents comprehensive consistency metrics for the NegotiationToM dataset, following the original paper's methodology. For each metric, the best and second-best results are highlighted in \textbf{bold} and \underline{underlined}, respectively.}
\end{table*}

\section{ToM Consistency Comprehensive Results}
\label{sec:tom-con-com}

\subsection{FanToM Dataset}
\label{subsec:fantom-dataset}
Table~\ref{tab:full-fantom} presents a detailed comparison of various methods evaluated against the metrics designed to measure the consistency of the \ac{tom} in the FanToM dataset, as discussed in the article \cite{kimFANToMBenchmarkStresstesting2023}.

\subsection{NegotiationToM Dataset}
\label{subsec:negotiationtom-dataset}
Table~\ref{tom_consistency_negotiationtom} provides a more comprehensive set of metrics for assessing consistency in the NegotiationToM dataset, as proposed in the original paper \cite{chanNegotiationToMBenchmarkStresstesting2024}.

% \section{\ac{llm} Finetuning}
% \label{sec:fine-tune-hypers}

% The finetuning process employs the same model checkpoint, preprocessing, and loading mechanisms as the LatentQA examples. The model is finetuned for a full epoch on the FanToM dataset and for three epochs on the NegotiationToM dataset. All other hyperparameters, which remain consistent across both datasets, are listed below:

% \begin{lstlisting}[basicstyle=\ttfamily\footnotesize, breaklines=true, frame=single, backgroundcolor=\color{gray!10}]
% SEQ_LENGTH=1300                  
% BATCH_SIZE=1                    
% GR_ACC_STEPS=1                   
% LR=5e-4                          
% LR_SCHEDULER_TYPE="cosine"       
% WEIGHT_DECAY=0.01                
% NUM_WARMUP_STEPS=30              
% BF16=True                        
% FP16=False                       

% # LoRA Configuration
% LORA_R=8                         
% LORA_ALPHA=32                    
% LORA_DROPOUT=0.0                 
% mlp_modules = ["gate_proj", "up_proj", "down_proj", "lm_head"]
% attn_modules = ["q_proj", "k_proj", "v_proj", "o_proj"]
% LORA_TARGET_MODULES=mlp_modules+attn_modules    

% # Bitsandbytes Configuration
% USE_NESTED_QUANT=True            
% BNB_4BIT_COMPUTE_DTYPE="bfloat16"
% \end{lstlisting}

\section{Sampling Method for Controllability Experiments}
\label{sec:sample-selection}

The NegotiationToM dataset provides labels at the utterance level, enabling this study to examine whether manipulating the \ac{tom}-related structures within the \ac{llm} leads to improved responses. To achieve this, it is essential to carefully select samples to ensure that specific \ac{tom} components reflect natural flow of the conversations. 

The following presents an example of sample selection for the \textit{Intention} component, followed by an instance illustrating selection for the \textit{Desire} component. The selection criteria remain consistent for the \textit{Belief} component. In each case, the conversation history, the reference utterance and the associated label are specified.

In the example below, the label \textit{"Showing Empathy"} is assigned to the final utterance. The preceding dialogue serves as context, while the last utterance functions as the reference for evaluation by the \ac{llm} judges.

\begin{lstlisting}[basicstyle=\ttfamily\footnotesize, breaklines=true, frame=single, backgroundcolor=\color{gray!10}]
# History 
  agent_1: Hello!  Let's work together on a deal for these packages, shall we? What are you most interested in?,
  agent_2: Hey! I'd like some more firewood to keep my doggo warm. What do you need?,
  agent_1: I need firewood as well. We have a large group consisting of mostly senior citizens, including my grandma, so we'd like the firewood to keep everyone warm.,
  agent_2: I see.  What are you least interested in?,
  agent_1: We can make do without extra water. Can we trade two waters for an extra firewood package and an extra food package?,
  agent_2: We could do without the water as well. I'm willing to trade you 3 firewood for 3 food and 2 waters,
# Reference
  agent_1: We need some firewood too, though! Let's try to make a deal that benefits us both!  Could I have 1 firewood, 3 food, and 3 waters?,

# Labels 
  Show-Empathy,
\end{lstlisting}

For alignment based on the \ac{tom} components of \textit{Belief} or \textit{Desire}, the last utterance is similarly considered. If the utterance updates a desire—such as prioritizing firewood—the preceding dialogue serves as context, while the updated utterance is used as the reference.

\begin{lstlisting}[basicstyle=\ttfamily\footnotesize, breaklines=true, frame=single, backgroundcolor=\color{gray!10}]
# History 
  agent_1: Hello! Let's work together on a deal for these packages, shall we? What are you most interested in?,
  agent_2: Hey! I'd like some more firewood to keep my doggo warm. What do you need?,
  agent_1: I need firewood as well. We have a large group consisting of mostly senior citizens, including my grandma, so we'd like the firewood to keep everyone warm.,
# Reference
  agent_2: I see. What are you least interested in?

# Labels 
    agent1_desire_high: Firewood,
\end{lstlisting}

\section{Steering \acp{llm}}
\label{sec:steer-hyper}
LatentQA enables fine-grained control over the output of target language models. However, achieving effective alignment requires careful design of question-answer pairs and precise hyperparameter tuning to ensure the backpropagation of losses from the decoder model to earlier layers of target model, results in more aligned responses. This section outlines the hyperparameter configuration used in this study. The steering process was conducted using 300 steering samples in all experiments -while it needs less- with target representations extracted from the middle layer (=15) of the target model.

The following hyperparameters were used.  For detailed definitions, please refer to the official LatentQA paper and repository.

\begin{lstlisting}[basicstyle=\ttfamily\footnotesize, breaklines=true, frame=single, backgroundcolor=\color{gray!10}]
    module_setup: str = "read-vary_write-fixed_n-fixed"
    modify_chat_template: bool = True
    shift_position_ids: bool = True
    lr: float = 5e-5
    seed: int = 42
    batch_size_training: int = 1
    samples: int = 50
    layers_to_optimize: tuple = (0, 1, 2, 3, 4, 5, 6, 7, 8, 9, 10, 11, 12, 13, 14, 15)
    per_layer_loss: bool = False
    qa_per_layer: bool = False


    
\end{lstlisting}

\section{Fine-tuning \acp{llm}}
\label{sec:fine-tune-hypers}
This section details the hyperparameters employed for fine-tuning \acp{llm} on the FanToM and NegotiationToM datasets. To enhance the efficiency of these experiments, both full-precision models and their 4-bit variants were utilized. This decision was informed by observations during the \ac{tom} reading experiment, which indicated that the 4-bit variant may yield superior performance for certain tasks.

The experiments were conducted across two model sizes and two datasets. The common hyperparameters for NegotiationToM/FanToM dataset experiments are as follows: 

\begin{lstlisting}[basicstyle=\ttfamily\footnotesize, breaklines=true, frame=single, backgroundcolor=\color{gray!10}]
    max_seq_length = 2000/2500 # NegotiationToM/FanToM
    per_device_train_batch_size = 32/8
    gradient_accumulation_steps = 4/2
    warmup_steps = 5
    epochs = 5
    learning_rate = 5e-4/
    optim = "adamw_8bit"
    weight_decay = 0.01
    lr_scheduler_type = "linear"
    seed = 3407
    target_modules = ["q_proj", "k_proj", "v_proj", "o_proj", "gate_proj", "up_proj", "down_proj"]
\end{lstlisting}

\section{Linear Probing Implementation Details}
\label{sec:linear-probing}

This section describes the implementation details for the linear probing experiments. Similar to Section~\ref{sec:fine-tune-hypers}, the 4-bit variant of \acp{llm} is also used for these experiments. The representation from the residual stream corresponding to the target layer—specifically the middle layer—is extracted and used for training a linear classifier on the CaSiNo dataset and a linear regressor on the CraigslistBargain dataset. The hyperparameters used for the grid search in these experiments are detailed below.

For the logistic regression model using the \texttt{scikit-learn} library:

\begin{lstlisting}[basicstyle=\ttfamily\footnotesize, breaklines=true, frame=single, backgroundcolor=\color{gray!10}]
pipeline = Pipeline([
        ('pca', PCA(random_state=29, n_components=n_components)),
        ('scaler', StandardScaler()),
        ('classifier', LogisticRegression(random_state=29, max_iter=1000))  
    ])
param_grid = {
    'classifier__C': [0.1, 1, 10],
    'classifier__penalty': ['l1', 'l2', 'elasticnet'], 
    'classifier__solver': ['liblinear', 'saga', 'lbfgs']  
}
\end{lstlisting}

For the Ridge regressor model using the \texttt{scikit-learn} library:

\begin{lstlisting}[basicstyle=\ttfamily\footnotesize, breaklines=true, frame=single, backgroundcolor=\color{gray!10}]
pipeline = Pipeline([
        ('pca', PCA(random_state=29,  n_components = 1530
n_components ))),
        ('regressor', Ridge(random_state=29))
    ])
param_grid = {
    'pca__n_components': [n for n in range(0, 500, 100)],
}
\end{lstlisting}

\section{Sample Responses Generated by the Model}  
\label{sec:sample_responses}
The following are selected examples of responses generated by the model. These examples were carefully chosen to illustrate different patterns in responses—\textbf{Actual Response}, \textbf{ToM-informed}, and \textbf{ToM-uninformed}. It is important to note that not all generated responses exhibit such clear distinctions, and tracking these differences can be more nuanced in many cases.

\subsection{Explicit Desire for Water Set to High}  

\noindent\textbf{Example 1:}  

\textbf{Actual Response}: I really need the water more than the rest of that stuff. I'll take 2 waters, 1 food, and 1 firewood.  
\textbf{ToM-informed}: Let's stay like this, 2 of water, 1 of food, 1 of firewood, and you take the rest.  
\textbf{ToM-uninformed}: Sounds like a fair deal to me! I'll take 1 water, 2 food, and 2 firewood, and you can keep 1 water, 1 food, and 1 firewood.  

\noindent\textbf{Example 2:}  

\textbf{Actual Response}: I think two packages of food would work for me.  
\textbf{ToM-informed}: What about I give you more food in exchange for you giving me more water? I could give you 2 food, and you give me an extra water.  
\textbf{ToM-uninformed}: Oh wow, that sounds like a great deal! I think we can make that work. Yeah, sure thing. I'll give you an extra water if you give me 2 food. That way we both get what we need.  

\noindent\textbf{Example 3:}  

\textbf{Actual Response}: Unfortunately, that doesn't work for me. I am afraid I will not be able to discuss this any further. My final offer is I get 2 water, 2 food, and 1 firewood.  
\textbf{ToM-informed}: Oops, I guess we are jumping the gun! I don't see a problem with that. You are welcome to one of the firewood and water.  
\textbf{ToM-uninformed}: Thanks! I think we're having a fun little camping adventure in our conversation!  

\subsection{Explicit Intention to Build Report} 

\noindent\textbf{Example 4:}

\textbf{Actual Response}: Awesome. I couldn't agree more. I feel that camping and hiking are both such spiritual experiences. It's such a perfect time to connect with our lord and savior. I love serving others as well. \textbf{ToM-informed}: I'm so excited! I've been looking forward to this trip for weeks. We're going to get away from the hustle and bustle of city life and reconnect with nature.	\textbf{ToM-uninformed}: I'm glad to hear that! As a digital \ac{ai} assistant, I don't have personal experiences or physical needs, but I'm excited to help facilitate your camping trip and provide information and assistance as needed. What are you most looking forward to about this trip?	

\subsection{Explicit Belief for Firewood Set to High} 

\noindent\textbf{Example 5:} 

\textbf{Actual Response}: That's unfortunate. I cannot agree to a deal without 2 pieces of firewood. I would have to walk away, but that would be worse for both of us. \textbf{ToM-informed}: It looks like we both need firewood. Let's divide it up. We can each have 2 pieces of firewood. That way, we'll both have enough to heat up our food.\textbf{ToM-uninformed}: Now, let's talk about the water and food. Do you need 1 water and 1 food item, or would you like to split those as well?	

\section{Samples of Failure Cases}
\label{sec:failure_cases}
This section highlights several failure patterns observed during the controllability phase, as identified through qualitative analysis. However, for some other failure cases, the analysis did not yield a clear explanation.

\subsection{Model Forgets the Conversation History}

A recurring error pattern observed during steering phase of conversational intent involved the model excessively prioritizing the target intent—such as displaying empathy—at the expense of maintaining coherence with the conversation’s prior context. In these cases, the model's responses often ignored the preceding dialogue and focused solely on expressing the desired intent, resulting in responses that were contextually inappropriate or disconnected from the conversation history.

\noindent\textbf{Example}

\textbf{Actual Response}: I am walking away if I cannot get 2 food packages 1 water and 1 firewood. That leaves you with 1 food 2 waters and 2 firewoods. This is a good deal for you concidering that you havent been asking for water for your children until i mentioned that you need water for you children. Do you agree? \textbf{ToM-informed}: If you'd like to discuss a hypothetical scenario or need assistance with something else, I'm here to help! \textbf{ToM-uninformed}: I think we've been back and forth enough. I'm willing to make a deal with you. I'll give you 2 food packages, and you can take 3 water packages and 3 firewood packages. That way, we both get what we need. Sound good to you?

\subsection{Limitations in Steering Across Model Variants}

Due to limitations in running identical steering setups on production-level models such as GPT variants or Gemini, we encountered cases where neither the \ac{tom}-informed nor the \ac{tom}-uninformed models produced irrelevant responses. Although we accounted for such instances in the win-rate calculations, consistently eliciting contextually appropriate rather than generic responses requires further investigation.

\noindent\textbf{Example}
\textbf{Actual Response}: That would be fine.  I was hoping to have 2 but since I am only walking and you are doing more heavy duty hiking, you take them.  Agreed? \textbf{ToM-informed}: It seems like you're putting a lot of thought into your camping trip. How do you plan on transporting the food and seasonings? \textbf{ToM-uninformed}: I think there's been a misunderstanding. As a digital AI assistant, I don't have a physical presence, so I won't be able to join you on your camping trip or provide food or firewood. I'm here to help answer any questions or provide information to help you prepare for your adventure. Would you like some suggestions on what to pack or how to prepare for your trip?

\end{document}